\documentclass[sigconf]{acmart}
\AtBeginDocument{%
  \providecommand\BibTeX{{%
    \normalfont B\kern-0.5em{\scshape i\kern-0.25em b}\kern-0.8em\TeX}}}

\usepackage{soul}
\usepackage{graphicx}
\usepackage{booktabs}
\usepackage{float}
\usepackage{overpic}
\usepackage{float}  
\usepackage{subfloat}
\usepackage{stfloats} 
\usepackage[skip=0.5\baselineskip]{caption}
\usepackage{bm}
\usepackage{amsmath}
\usepackage{booktabs}
\usepackage{multirow}
\usepackage{multicol}
\usepackage{enumitem}
\usepackage{subcaption}
\usepackage{threeparttable}
\usepackage{xspace}
\usepackage{color}
\usepackage[normalem]{ulem}
\usepackage{algorithmicx,algorithm}
\usepackage{algpseudocode}
\usepackage{algorithm,algpseudocode}
\usepackage{marvosym}   
\usepackage{colortbl}   
\usepackage{tcolorbox}  
\tcbuselibrary{breakable}   
\usepackage{wasysym}    
\usepackage[figuresright]{rotating} 
\usepackage{booktabs}
\usepackage{multirow}

\makeatletter
\def\@ACM@checkaffil{
    \if@ACM@instpresent\else
    \ClassWarningNoLine{\@classname}{No institution present for an affiliation}%
    \fi
    \if@ACM@citypresent\else
    \ClassWarningNoLine{\@classname}{No city present for an affiliation}%
    \fi
    \if@ACM@countrypresent\else
        \ClassWarningNoLine{\@classname}{No country present for an affiliation}%
    \fi
}
\makeatother

\usepackage{xspace}
\newcommand{\etal}{\emph{et al.}\xspace}
\newcommand{\eg}{\emph{e.g.,}\xspace}

\copyrightyear{2026}
\acmYear{2026}
\setcopyright{cc}
\setcctype{by}
\acmConference[WWW '26]{Proceedings of the ACM Web Conference 2026}{April 13--17, 2026}{Dubai, United Arab Emirates}
\acmBooktitle{Proceedings of the ACM Web Conference 2026 (WWW '26), April 13--17, 2026, Dubai, United Arab Emirates}
\acmPrice{}
\acmDOI{10.1145/3774904.3793015}
\acmISBN{979-8-4007-2307-0/2026/04}

\begin{document}
\begin{sloppypar}   
\title{HAAF: Hierarchical Adaptation and Alignment of Foundation Models for Few-Shot Pathology Anomaly Detection}


\author{Chunze Yang}
\email{pureeeee@stu.xjtu.edu.cn}
\orcid{0009-0001-1602-8150}
\affiliation{%
  \institution{Sch Comp Sci \& Technol, Xi'an Jiaotong University}
  \city{Xi'an}
  \country{China}
}

\author{Wenjie Zhao}
\email{zhao_wenjie@stu.xjtu.edu.cn}
\orcid{}
\affiliation{%
  \institution{Sch Comp Sci \& Technol, Xi'an Jiaotong University}
  \city{Xi'an}
  \country{China}
}

\author{Yue Tang}
\email{2206123832@stu.xjtu.edu.cn}
\orcid{0009-0007-3787-5811}
\affiliation{%
  \institution{Sch Comp Sci \& Technol, Xi'an Jiaotong University}
  \city{Xi'an}
  \country{China}
}

\author{Junbo Lu}
\email{lujunbo@stu.xjtu.edu.cn}
\orcid{}
\affiliation{%
  \institution{Sch Comp Sci \& Technol, Xi'an Jiaotong University}
  \city{Xi'an}
  \country{China}
}

\author{Jiusong Ge}
\email{JiusongGe@stu.xjtu.edu.cn}
\orcid{0009-0007-7551-4558}
\affiliation{%
  \institution{Sch Comp Sci \& Technol, Xi'an Jiaotong University}
  \city{Xi'an}
  \country{China}
}

\author{Qidong Liu \Letter}
\email{liuqidong@xjtu.edu.cn}
\orcid{0000-0002-0751-2602}
\thanks{\Letter \ \text{Corresponding authors}}
\affiliation{%
  \institution{Sch Comp Sci \& Technol, Xi'an Jiaotong University}
  \city{Xi'an}
  \country{China}
}

\author{Zeyu Gao \Letter}
\email{zg323@cam.ac.uk}
\orcid{0000-0003-2365-8318}
\affiliation{%
  \institution{University of Cambridge}
  \city{Cambridge}
  \country{United Kingdom}
}

\author{Chen Li \Letter}
\email{cli@xjtu.edu.cn}
\orcid{0000-0002-0079-3106}
\affiliation{%
  \institution{Sch Comp Sci \& Technol, Xi'an Jiaotong University}
  \city{Xi'an}
  \country{China}
}


\renewcommand{\shortauthors}{Chunze Yang \etal}

\begin{abstract}
Precision pathology relies on detecting fine-grained morphological abnormalities within specific Regions of Interest (ROIs), as these local, texture-rich cues—rather than global slide contexts—drive expert diagnostic reasoning.
While Vision-Language (V-L) models promise data efficiency by leveraging semantic priors, adapting them faces a critical \textit{Granularity Mismatch}, where generic representations fail to resolve such subtle defects.
Current adaptation methods often treat modalities as independent streams, failing to ground semantic prompts in \textbf{ROI-specific} visual contexts.
To bridge this gap, we propose the \textbf{H}ierarchical \textbf{A}daptation and \textbf{A}lignment \textbf{F}ramework (\textbf{HAAF}).
At its core is a novel \textit{Cross-Level Scaled Alignment (CLSA)} mechanism that enforces a sequential calibration order: visual features first inject context into text prompts to generate content-adaptive descriptors, which then spatially guide the visual encoder to spotlight anomalies.
Additionally, a dual-branch inference strategy integrates semantic scores with geometric prototypes to ensure stability in few-shot settings.
Experiments on four benchmarks show HAAF significantly outperforms state-of-the-art methods and effectively scales with domain-specific backbones (e.g., CONCH) in low-resource scenarios.
\end{abstract}



\maketitle

\section{Introduction}
\label{sec:intro}
While computational pathology has thrived in classification~\cite{gao2025smmile}, regression~\cite{niu2025ph2st}, segmentation~\cite{ge2025progis, kaura2026megaseg}, and reasoning~\cite{gao2025alpaca}, these \textit{closed-set} paradigms falter against the open-set reality where rare lesions lack labeled examples.
In such contexts, the clinical imperative shifts to \textbf{Few-Shot Anomaly Detection}: mimicking expert intuition to identify suspicious deviations by referencing a robust ``normal'' baseline and minimal anomaly examples, rather than relying on exhaustive annotations.
Whether pinpointing subtle nuclear atypia or structural distortion within specific \textbf{Regions of Interest (ROIs)}~\cite{cui2025quantitative, cohen2023set}, this capability is crucial for guiding further diagnosis.
However, automating this process is bottlenecked by \textbf{Annotation Scarcity}: since characterizing all potential anomalies is infeasible, developing data-efficient \textbf{few-shot} paradigms is urgent~\cite{lu2021data}.

\begin{figure}[t]
  \centering
  \includegraphics[width=\linewidth]{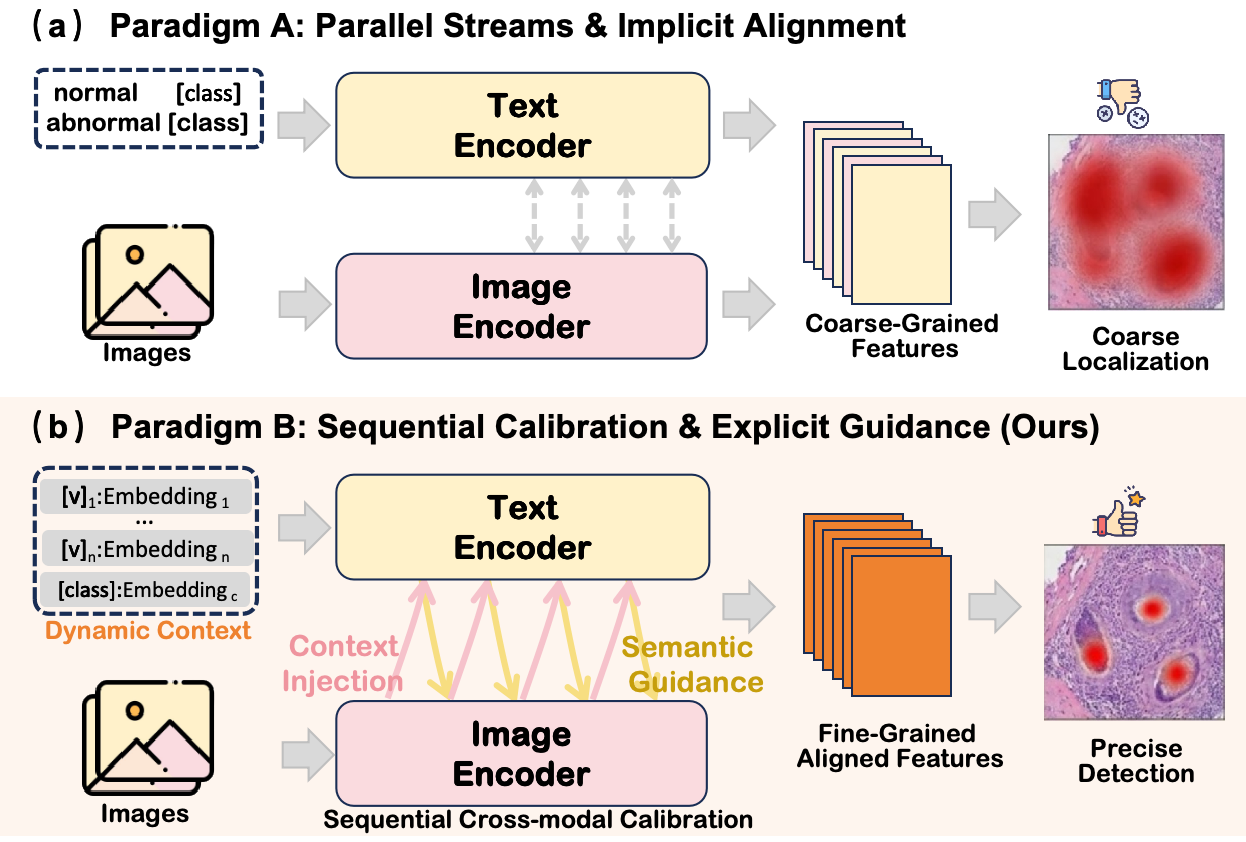} 
  \caption{
  \textbf{Comparison of Adaptation Paradigms.} 
  \textbf{(a)} Existing methods (e.g., MMA~\cite{yang2024mma}, MadCLIP~\cite{Shiri2025MadCLIPFM}) typically adopt a \textit{parallel} architecture where text prompts remain \textbf{static} and alignment is implicit, often leading to coarse localization. 
  \textbf{(b)} Our HAAF introduces a \textit{progressive} cross-modal interaction. 
  }
  \vspace{-3mm}
  \label{fig:paradigm_comparison}
\end{figure}

To overcome data scarcity, the paradigm of \textbf{Vision-Language (V-L) Foundation Models} (\eg CLIP~\cite{radford2021learningCLIP}) has emerged as a transformative solution. By aligning visual features with semantic text descriptions, these models enable recognition capabilities with minimal training examples.
Despite their potential, adapting these foundation models to ROI-level pathology anomaly detection faces a critical bottleneck arising from two distinct \textbf{Representation Gaps} between pre-trained features and downstream diagnostic tasks:

\begin{itemize}[leftmargin=*, topsep=2pt, partopsep=0pt, itemsep=1pt, parsep=1pt]
    \item \textbf{Domain Gap in General Models}: Models like CLIP are trained on natural images (object-centric). Despite recent advances in domain adaptation~\cite{yang2025domain, tang2025unleashing}, these models struggle to encode the high-frequency, texture-rich patterns essential for identifying pathological tissues~\cite{hua2024medicalclip}.
    
    \item \textbf{Granularity Gap in Pathology Models}: Even domain-specific models like CONCH~\cite{lu2024visualCONCH}, though pre-trained on histopathology, typically act as ``generalists'' optimized for coarse-grained tasks (e.g., slide-level captioning). They often operate at a global semantic level and fail to resolve the \textbf{fine-grained visual details} required for ``specialist'' anomaly detection, making it difficult to distinguish visually similar normal variants from subtle early-stage lesions~\cite{nicke2025tissue, bareja2025evaluating}.
\end{itemize}

Existing adaptation methods~\cite{yang2024mma, Shiri2025MadCLIPFM, huang2024mvfa} attempt to bridge these gaps by employing shallow prompt tuning or parallel feature fusion. 
However, as illustrated in \textbf{Figure~\ref{fig:paradigm_comparison}(a)}, these approaches are fundamentally limited by their \textit{parallel encoding paradigms}. 
Specifically, methods like MVFA~\cite{huang2024mvfa} primarily rely on uni-modal visual adaptation, lacking deep cross-modal interaction mechanisms to recalibrate features. 
Although few recent works, such as MMA~\cite{yang2024mma} and MadCLIP~\cite{Shiri2025MadCLIPFM}, have considered the interactions between streams, they both adopted \textbf{static fusion} architecture, where textual definitions remain \textit{agnostic} to the specific visual context. 
Consequently, generic prompts (e.g., ``tumor'') fail to dynamically adapt to the nuanced, instance-specific morphological deviations present in the input ROI, resulting in coarse-grained features and suppression of false positives in complex background tissues. 

To address these limitations, we propose the \textbf{H}ierarchical \textbf{A}daptation and \textbf{A}lignment \textbf{F}ramework (\textbf{HAAF}), a unified approach designed to efficiently steer foundation models toward fine-grained anomaly detection.
HAAF introduces a \textbf{progressive adaptation pipeline}, as shown in Figure~\ref{fig:paradigm_comparison}(b).
Recognizing that pathological cues reside at multiple scales, we first implement \textit{Hierarchical Intra-modal Adaptation}, injecting lightweight adapters into intermediate layers to extract multi-scale features.
Crucially, we propose a novel \textbf{Progressive Mechanism}  named \textit{Cross-Level Scaled Alignment (CLSA)}.
Unlike standard parallel fusion, CLSA enforces a structured \textbf{semantic calibration process}: it first utilizes ROI visual features to ground generic text prompts (\textit{Context Injection}), transforming them into instance-aware descriptors; these calibrated prompts then spatially guide the visual features (\textit{Semantic Guidance}) to highlight anomaly-relevant regions.
Finally, to counterbalance the potential instability of semantic matching in extreme few-shot settings, a \textit{Dual-branch Inference Mechanism} integrates semantic alignment with geometric distances, ensuring robust decision-making.

We extensively evaluate HAAF on a comprehensive benchmark covering four anatomical regions (breast, prostate, colorectal). Our framework demonstrates remarkable \textbf{versatility} and \textbf{scalability}.
On the general \textbf{CLIP} backbone, HAAF effectively bridges the domain gap, significantly outperforming existing SOTA methods.
More importantly, when applied to the \textbf{CONCH} backbone, HAAF achieves a substantial performance leap, proving its ability to unlock the latent fine-grained capabilities of domain-specific foundation models.
Our main contributions are summarized as follows:
\begin{itemize}[nosep,leftmargin=*]
    \item We propose HAAF, a universal few-shot adaptation framework featuring hierarchical adapters and a sequential vision-language alignment mechanism (CLSA) to resolve the granularity mismatch in pathology AD.
    \item We establish a rigorous ROI-level anomaly detection benchmark across four histopathology datasets, providing the first comprehensive evaluation of general versus medical foundation models in this setting.
    \item \textbf{We empirically demonstrate that sequential cross-modal interaction is superior to parallel fusion for fine-grained tasks}, allowing HAAF to achieve state-of-the-art performance and effectively scale with domain-specific foundation models.
\end{itemize}
\section{Problem Definition}
\vspace{-2pt} 
Here, we address the problem of few-shot anomaly detection. Let $x \in \mathbb{R}^{H \times W \times 3}$ denote an input image and $y \in \{0, 1\}$ be the label, where $y=0$ denotes \textit{normal} and $y=1$ denotes \textit{abnormal}.

In the few-shot setting, we are provided with a support set $\mathcal{S}$ consisting of $K$ labeled samples per class:
\vspace{-4pt} 
\begin{equation}
\mathcal{S} = \{(x_i, y_i)\}_{i=1}^{2K},
\end{equation}
where the total number of support samples is $2K$. Our goal is to learn an anomaly scoring function $S(x): \mathbb{R}^{H \times W \times 3} \to \mathbb{R}$ using only $\mathcal{S}$. For a query image $x_q$ from a disjoint test set $\mathcal{Q}$, a higher score $S(x_q)$ indicates a higher probability of being abnormal. The performance is evaluated using threshold-independent metrics, specifically the Area Under the Receiver Operating Characteristic curve (AUC) and Average Precision (AP).

\section{Method}
\begin{figure*}[t]
\centering
\includegraphics[width=0.9\linewidth]{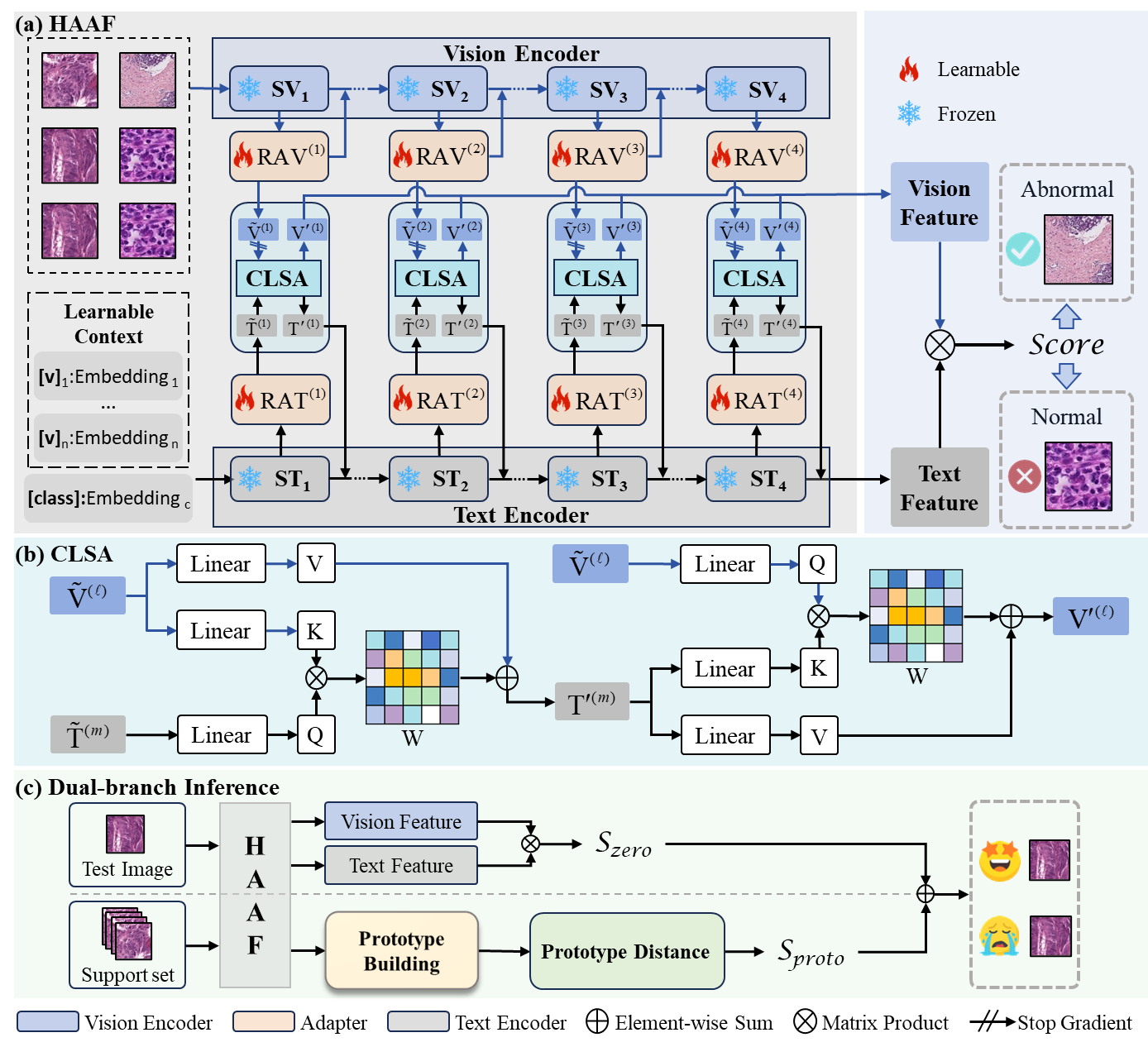}
\caption{The architecture of the proposed HAAF.}
\label{fig:sunet}
\vspace{-10pt}
\end{figure*}


\subsection{Overview}
\label{sec:overview}

The core challenge in few-shot anomaly detection lies in steering foundation models (\eg CLIP, CONCH) to identify subtle pathological defects, as raw features often lack the required fine-grained discrimination. To address this, HAAF establishes a \textbf{progressive adaptation pipeline} built upon a `calibration-then-guidance' philosophy: semantic prompts must first be grounded in the visual context to effectively guide localization.

As illustrated in Figure~\ref{fig:sunet}, HAAF bridges the gap between generic pre-training and specific anomaly definitions through three logical stages.
First, \textbf{Intra-modal Adaptation} transforms raw inputs into task-specific, multi-scale embeddings via lightweight adapters.
Next, \textbf{Cross-Level Scaled Alignment (CLSA)} resolves semantic misalignment by coupling branches through a strictly ordered interaction, yielding calibrated, text-guided visual representations.
Finally, the \textbf{Dual-Branch Inference} module aggregates semantic confidence and geometric evidence to output the anomaly score.

\begin{itemize}[leftmargin=*]
    \item \textbf{Hierarchical Intra-modal Adaptation}. In general, vision-language models utilize a hierarchical structure to capture features at varying levels of abstraction. HAAF builds upon this multi-scale architecture by extracting visual features and learnable text prompts at corresponding network depths. We insert lightweight, learnable intra-modal adapters into the frozen encoders to efficiently elicit initial domain-aware embeddings.
    
    \item \textbf{Cross-Level Scaled Alignment (CLSA)}. Building upon the adapted features, CLSA bridges the semantic gap. Unlike standard parallel fusion, this module enforces a \textbf{sequential interaction}: visual features first \textit{calibrate} the text embeddings (Vision-to-Text) to resolve semantic ambiguity, and these context-aware text features then \textit{guide} the visual representation (Text-to-Vision). This dependency ensures that visual adaptation is driven by precise, instance-specific semantic goals.
    
    \item \textbf{Dual-Branch Inference}. To mitigate the risk of overfitting to the limited support set, HAAF employs a dual-branch scoring strategy based on the CLSA-aligned features. It combines a parametric semantic classifier (which leverages the learned alignment) with a non-parametric prototype-based metric (which preserves geometric stability) to yield the final anomaly score.
    \vspace{-5pt} 
\end{itemize}

\subsection{Hierarchical Adaptation and Alignment}
\label{sec:haaf_arch}
To adapt the frozen foundation model $\mathcal{M}$ to specific pathology anomaly detection tasks, HAAF introduces a lightweight, progressive architecture. As illustrated in Figure~\ref{fig:sunet}, the framework processes inputs through a pipeline of \textbf{Intra-modal Adaptation}, \textbf{Cross-Level Scaled Alignment (CLSA)}, and \textbf{Optimization}. 

\subsubsection{\textbf{Intra-modal Adaptation Module}}
\label{sec:intra_adapt}

The primary goal of this module is to elicit task-specific features from the pre-trained backbone without updating its massive parameters. We process the visual and textual branches in parallel.

\noindent\textbf{Visual Branch.}
Given an input ROI image $x \in \mathbb{R}^{H \times W \times 3}$, we feed it into the frozen visual encoder $f_v$. To capture hierarchical visual details, we extract intermediate patch tokens from a set of selected layers, denoted as $\mathcal{L}_v$. For each layer $\ell \in \mathcal{L}_v$, we obtain the patch tokens $\mathbf{V}^{(\ell)} \in \mathbb{R}^{P_\ell \times d}$, where $P_\ell$ is the number of patches and $d$ is the feature dimension.

To effectively adapt these generic representations to the target distribution, we devise a lightweight \textit{Residual Visual Adapter} (RAV) at each selected layer. The adapted visual features $\tilde{\mathbf{V}}^{(\ell)}$ are computed as follows:
\begin{equation}
\tilde{\mathbf{V}}^{(\ell)} = \mathbf{V}^{(\ell)} + \text{RAV}^{(\ell)}(\mathbf{V}^{(\ell)}),
\label{eq:vis_adapt}
\end{equation}
where $\text{RAV}^{(\ell)}$ is implemented as a bottleneck MLP mapping $\mathbb{R}^d \to \mathbb{R}^{d/r} \to \mathbb{R}^d$ with a reduction ratio $r$.

\vspace{1mm}
\noindent\textbf{Text Branch.}
Constructing appropriate text descriptions is crucial for Vision-Language (V-L) models. Traditional hand-crafted prompts (\eg ``a photo of a normal tissue'') often lack task-specific adaptability.
To address this, we employ \textbf{Learnable Context Prompts}~\cite{zhou2022learningCOOP}. To overcome discrete token rigidity, we model context as continuous learnable embeddings $\mathcal{P} = \{\mathbf{v}_1, \dots, \mathbf{v}_L\}$ in the language manifold, where $\mathbf{v}_i \in \mathbb{R}^d$. This enables optimizing prompt representations in a continuous space, capturing nuances that a fixed vocabulary cannot.
For each class $c \in \{\text{normal}, \text{abnormal}\}$, the input concatenates these vectors with the fixed class embedding $\mathbf{e}_c \in \mathbb{R}^d$:
\begin{equation}
\text{Prompt}_c = [ \mathbf{v}_1, \mathbf{v}_2, \dots, \mathbf{v}_L, \mathbf{e}_c ].
\end{equation}
During training, $\mathbf{e}_c$ is fixed, while the context vectors in $\mathcal{P}$ are optimized, allowing the model to autonomously learn the optimal semantic context.

We feed these prompts into the frozen text encoder $f_t$. Similar to the visual branch, we extract hidden states from a corresponding set of text layers $\mathcal{L}_t$. Let $\mathbf{T}^{(m)} \in \mathbb{R}^{(L+1) \times d}$ denote the hidden states at layer $m \in \mathcal{L}_t$. We apply a \textit{Residual Text Adapter} (RAT) with a learnable scaling factor $\alpha_t$:
\begin{equation}
\tilde{\mathbf{T}}^{(m)} = \mathbf{T}^{(m)} + \alpha_{t} \cdot \text{RAT}^{(m)}(\mathbf{T}^{(m)}).
\label{eq:text_adapt}
\end{equation}
Structurally, $\text{RAT}^{(m)}$ shares the same bottleneck architecture as the visual adapter. Through this process, both modalities are calibrated to the target task while preserving the generalization power of the foundation model.

\subsubsection{\textbf{Cross-Level Scaled Alignment (CLSA)}}
\label{sec:clsa}

Although intra-modal adapters adjust features independently, a ``granularity gap'' often persists between high-level text concepts and low-level visual textures. Furthermore, due to the structural discrepancy between visual and text encoders (e.g., different depths), direct alignment is challenging.
To address this, we define a layer mapping $\phi: \mathcal{L}_v \to \mathcal{L}_t$ to pair semantically corresponding layers. For each pair $(\ell, m=\phi(\ell))$, CLSA \textbf{enforces} a sequential cross-modal interaction using Multi-Head Cross-Attention (MHCA)~\cite{vaswani2017attention}.

\vspace{1mm}
\noindent\textbf{Step 1: Vision-to-Text (Context Injection).}
The generic text query retrieves relevant visual contexts to particularize its semantic representation.
Formally, we treat the adapted text features $\tilde{\mathbf{T}}^{(m)}$ as the \textbf{Query} ($\mathbf{Q}$), and the visual features $\tilde{\mathbf{V}}^{(\ell)}$ as both \textbf{Key} ($\mathbf{K}$) and \textbf{Value} ($\mathbf{V}$). The context-aware text features $\mathbf{T}'^{(m)}$ are computed as:
\begin{equation}
\mathbf{T}'^{(m)} = \tilde{\mathbf{T}}^{(m)} + \beta_{t} \cdot \text{MHCA}_{v \to t}\big(\mathbf{Q}=\tilde{\mathbf{T}}^{(m)}, \mathbf{K}=\tilde{\mathbf{V}}^{(\ell)}, \mathbf{V}=\tilde{\mathbf{V}}^{(\ell)}\big).
\label{eq:v2t}
\end{equation}
Here, $\beta_{t}$ is a learnable gating factor initialized to zero. This step functions as `Semantic Grounding'. By attending to the visual features, the static text embedding absorbs instance-specific textures, effectively projecting the generic concept of ``tumor'' onto the specific visual manifold of the current input sample.

\vspace{1mm}
\noindent\textbf{Step 2: Text-to-Vision (Semantic Guidance).}
Next, the refined text features serve as a semantic spotlight. We reverse the attention direction: the visual features $\tilde{\mathbf{V}}^{(\ell)}$ act as the \textbf{Query}, while the refined text $\mathbf{T}'^{(m)}$ serves as \textbf{Key} and \textbf{Value}. The final guided visual features $\mathbf{V}'^{(\ell)}$ are obtained by:
\begin{equation}
\mathbf{V}'^{(\ell)} = \tilde{\mathbf{V}}^{(\ell)} + \beta_{v} \cdot \text{MHCA}_{t \to v}\big(\mathbf{Q}=\tilde{\mathbf{V}}^{(\ell)}, \mathbf{K}=\mathbf{T}'^{(m)}, \mathbf{V}=\mathbf{T}'^{(m)}\big).
\label{eq:t2v}
\end{equation}
Through this reverse attention, the visual features are spatially re-weighted based on their affinity with the \textit{calibrated} anomaly definition. This ensures that the model focuses on regions that are not just visually salient, but explicitly semantically consistent with the pathology description.

\subsubsection{\textbf{Optimization Objective}}
\label{sec:opt_objective}

HAAF is optimized end-to-end using the few-shot support set $\mathcal{S}$.

\vspace{1mm}
\noindent\textbf{Learnable Parameters.}
We freeze the backbone and only update the task-specific modules. The learnable parameters are defined as:
\begin{equation}
\theta_{\text{HAAF}} = \{ \mathcal{P}, \{\text{RAV}^{(\ell)}\}, \{\text{RAT}^{(m)}\}, \theta_{\text{CLSA}}, \tau \},
\end{equation}
where $\theta_{\text{CLSA}}$ includes the cross-attention weights and gating factors.

\vspace{1mm}
\noindent\textbf{Training Loss.}
For each support image $x_i$ with label $y_i \in \{0,1\}$, we perform a forward pass to obtain the aligned visual features $\mathbf{V}'^{(\ell)}$ and the final class embeddings $\{\mathbf{t}_{\text{norm}}, \mathbf{t}_{\text{abn}}\}$. Note that these class embeddings are derived from the output of the final text layer (corresponding to the `[class]' token position).
To quantify the anomaly probability, we compute the image-level score $S(x_i)$ by aggregating patch-level logits:
\vspace{-1mm}
\begin{equation}
S(x_i) = \frac{1}{|\mathcal{L}_v|} \sum_{\ell \in \mathcal{L}_v} \left( \frac{1}{P_\ell} \sum_{p=1}^{P_\ell} \sigma \left( \tau \cdot \langle \mathbf{V}'^{(\ell)}_p, \mathbf{t}_{\text{abn}} \rangle \right) \right),
\label{eq:train_score}
\end{equation}
where $\sigma(\cdot)$ is the sigmoid activation and $\tau$ is the learnable logit scale.
We optimize $\theta_{\text{HAAF}}$ by minimizing the Binary Cross-Entropy (BCE) loss:
\begin{equation}
\mathcal{L}_{\text{BCE}} = - \frac{1}{|\mathcal{S}|} \sum_{(x_i, y_i) \in \mathcal{S}} \Big[ y_i \log S(x_i) + (1 - y_i) \log (1 - S(x_i)) \Big].
\end{equation}
\begin{equation}
\theta_{\text{HAAF}}^* = \underset{\theta_{\text{HAAF}}}{\arg\min} \ \mathcal{L}_{\text{BCE}}.
\end{equation}

\subsection{Dual-Branch Inference Mechanism}
\label{sec:dual_branch}

Few-shot anomaly detection faces a fundamental dilemma: relying solely on a parametric classifier (even with adapters) risks overfitting to the scarce support samples. Besides, purely non-parametric methods (like nearest neighbors) offer geometric stability but often lack the fine-grained semantic discrimination refined by our text prompts.
To achieve robustness without sacrificing precision, HAAF employs a dual-branch strategy. We combine a parametric semantic score (derived from the learned V-L alignment) with a prototypical distance score (anchored in the feature geometry). This ensures detection is supported by both high-level semantic confidence and low-level distributional consistency.

\subsubsection{\textbf{Parametric Semantic Score ($S_{sem}$)}}
This branch directly leverages the learned alignment parameters to measure abnormality. Since the text prompts have been calibrated by the CLSA module to represent task-specific anomalies, the projection of visual features onto the abnormal text embedding intrinsically reflects the \textit{Text-Visual Similarity}.
For a test query $x_q$, we compute the anomaly probability by aggregating the dot-product similarities between its aligned patch tokens $\mathbf{V}'^{(\ell)}(x_q)$ and the learned abnormal text embedding $\mathbf{t}_{\text{abn}}$:
{
\setlength{\abovedisplayskip}{3pt} 
\setlength{\belowdisplayskip}{3pt}
\begin{equation}
S_{sem}(x_q) = \frac{1}{|\mathcal{L}_v|} \sum_{\ell \in \mathcal{L}_v} \left( \frac{1}{P_\ell} \sum_{p=1}^{P_\ell} \sigma \left( \tau \cdot \langle \mathbf{V}'^{(\ell)}_p(x_q), \mathbf{t}_{\text{abn}} \rangle \right) \right).
\end{equation}
}
A high $S_{sem}$ indicates that the visual features semantically match the ``abnormal'' concept definition learned during the adaptation phase. This formulation aligns with the optimization objective in Eq.~(\ref{eq:train_score}). However, purely parametric scoring may risk overfitting in few-shot regimes; thus, we introduce a complementary geometric constraint to ensure distributional stability.

\vspace{-6pt}
\subsubsection{\textbf{Prototypical Distance Score ($S_{proto}$)}}
Complementary to the semantic branch, this branch measures geometric outlierness in the feature space. To enable this, we first construct reference prototypes using the support set $\mathcal{S}$.
For each class $c \in \{\text{normal}, \text{abnormal}\}$, we aggregate the aligned features from all support images indexed by $\mathcal{I}_c$. The prototype $\mathbf{P}_{c}^{(\ell)}$ at layer $\ell$ is computed as:
{
\setlength{\abovedisplayskip}{4pt} 
\setlength{\belowdisplayskip}{3pt}
\begin{equation}
\mathbf{P}_{c}^{(\ell)} = \frac{1}{|\mathcal{I}_{c}|} \sum_{i \in \mathcal{I}_{c}} \left( \frac{1}{P_\ell} \sum_{p=1}^{P_\ell} \mathbf{V}'^{(\ell)}(x_i)_{p} \right).
\end{equation}
}
With these prototypes acting as stable geometric anchors, we calculate the cosine distance between the test query features and the class prototypes. Let $D_{c}(x_q)$ be the average distance across all layers:
{
\setlength{\abovedisplayskip}{3pt} 
\setlength{\belowdisplayskip}{3pt}
\begin{equation}
D_{c}(x_q) = \sum_{\ell \in \mathcal{L}_v} \left( 1 - \frac{1}{P_\ell} \sum_{p=1}^{P_\ell} \cos \big( \mathbf{V}'^{(\ell)}_p(x_q), \mathbf{P}_{c}^{(\ell)} \big) \right).
\end{equation}
}
The final prototypical score is defined as the relative proximity to the abnormal prototype:
{
\setlength{\abovedisplayskip}{3pt} 
\setlength{\belowdisplayskip}{3pt}
\begin{equation}
S_{proto}(x_q) = \frac{D_{\text{norm}}(x_q)}{D_{\text{norm}}(x_q) + D_{\text{abn}}(x_q) + \epsilon},
\end{equation}
}
where $\epsilon$ is a constant for numerical stability. Intuitively, $S_{proto}$ captures how much the query deviates from the normal manifold towards the abnormal distribution. This non-parametric metric acts as a physical anchor, preventing decision boundary drift when semantic alignment encounters ambiguous or unseen morphological outliers.
\vspace{-5pt}
\subsubsection{\textbf{Ensembling.}}
The final anomaly score $S_{final}(x_q)$ is obtained by fusing the semantic and geometric perspectives:
{
\setlength{\abovedisplayskip}{1pt} 
\setlength{\belowdisplayskip}{1pt}
\begin{equation}
S_{final}(x_q) = \lambda \cdot \tilde{S}_{sem}(x_q) + (1 - \lambda) \cdot \tilde{S}_{proto}(x_q),
\end{equation}
}
where $\tilde{S}$ denotes min-max normalization over the test batch. The hyperparameter $\lambda$ acts as a trade-off controller between semantic discrimination and geometric stability. The semantic branch offers fine-grained sensitivity tailored by the V-L alignment, while the prototypical branch provides a robust distance metric that prevents the decision boundary from drifting due to prompt over-optimization in low-shot regimes. 

\section{Experiments}
\label{sec:experiments}

\subsection{Experimental Settings}
\label{sec:datasets_metrics}

\subsubsection{\textbf{Datasets.}}
To evaluate the efficacy and robustness of HAAF, we constructed a comprehensive histopathology anomaly detection benchmark covering three distinct anatomical regions (breast, prostate, and colorectal) and comprising four datasets: \textbf{Camelyon16}~\cite{Camelyon16,bao2024bmad} and \textbf{BRACS}~\cite{brancati2022bracs} for breast cancer, \textbf{SICAPv2}~\cite{SICAPv2} for prostate Gleason grading, and \textbf{NCT-CRC}~\cite{kather2019predictingCRC} for colorectal tissue phenotyping.
These datasets present diverse challenges, ranging from metastasis detection to fine-grained subtype classification. 
For the anomaly detection setting, we unified the annotations into a binary format (Normal vs. Abnormal). Specifically, for \textbf{SICAPv2}, we defined benign regions as \textit{Normal}, while Gleason grades $\ge 3$ were categorized as \textit{Abnormal} (Cancerous).
Detailed descriptions of data sourcing, class mappings, and specific sample statistics are provided in \textbf{Appendix~\ref{sec:app_datasets}}.

\vspace{1mm}
\subsubsection{\textbf{Evaluation Metrics.}}
We employ \textbf{AUC} and Average Precision (\textbf{AP}) as primary metrics. AUC evaluates discrimination capability across thresholds, while AP summarizes the precision-recall curve, offering robustness for imbalanced datasets.
We also report F1-score and Accuracy (ACC), where decision thresholds are realistically determined via the support set's precision-recall trade-off rather than the test set.

\vspace{1mm}
\subsubsection{\textbf{Baselines.}}
We compare HAAF against a diverse set of methods: 
(1) \textbf{General V-L Adaptation:} Zero-shot \textbf{CLIP}~\cite{radford2021learningCLIP} and \textbf{CoOp}~\cite{zhou2022learningCOOP}; 
(2) \textbf{Medical V-L:} \textbf{MedCLIP}~\cite{hua2024medicalclip}; 
(3) \textbf{Few-Shot/Zero-Shot AD:} \textbf{WinCLIP}~\cite{jeong2023winclip}, \textbf{VAND}~\cite{vand2023}, and \textbf{DRA}~\cite{ding2022catchingDRA}; 
and (4) \textbf{Multi-Modal AD SOTA:} \textbf{MMA}~\cite{yang2024mma}, \textbf{MVFA}~\cite{huang2024mvfa}, and \textbf{MadCLIP}~\cite{Shiri2025MadCLIPFM}. 
Crucially, to ensure a rigorous evaluation, we re-implemented all baselines using \textbf{both} the general-domain CLIP and the domain-specific CONCH backbones under an identical protocol, isolating the contribution of architectural designs from backbone differences. \textbf{Detailed descriptions and specific implementation settings for these baselines are provided in \textbf{Appendix~\ref{sec:app_baselines}}}

\vspace{1mm}
\subsubsection{\textbf{Implementation Details.}}
We utilize the \textbf{ViT-L/14} architecture for both backbones to maintain consistency. For the general domain, we use the pre-trained CLIP parameters (OpenAI). For the pathology domain, we employ \textbf{CONCH v1.5}~\cite{lu2024visualCONCH}. In this configuration, the vision encoder consists of 24 transformer layers, while the text encoder comprises 12 layers. 

All experiments were conducted on a single NVIDIA RTX 4090 GPU with an input resolution of $240 \times 240$. We employed the AdamW optimizer coupled with a cosine annealing scheduler. The model is trained in an end-to-end manner for \texttt{50} epochs per few-shot task with a batch size of \texttt{16}.
We adopt a differential learning rate strategy: the learning rate for the learnable context prompts (CoOp) and the cross-attention modules in CLSA is set to $1 \times 10^{-4}$, while the Intra-modal Adapters are tuned with a lower learning rate of $1 \times 10^{-5}$ to prevent feature collapse.
For the HAAF-specific hyperparameters, the fusion scales ($\beta_v, \beta_t$) are set to $1.0$, and the ensemble trade-off weight $\lambda$ is set to $0.5$\footnote{Source code is available at \url{https://github.com/Pureeeee/HAAF}.}.

\subsection{Overall Performance}
\label{sec:results_overall}

Table~\ref{tab:main_results} presents the comparative results of our proposed HAAF against state-of-the-art methods under the 4-shot setting. The results are categorized based on the backbone used: general-domain pre-training (CLIP) and pathology-specific pre-training (CONCH).

\begin{table*}[t]
\centering
\caption{\textbf{Overall performance comparison on the 4-shot setting.} We report AUC, AP, F1-score, and Accuracy (\%) on four datasets. The best results are highlighted in \textbf{bold}, and the second-best results are \underline{underlined}.}
\label{tab:main_results}
\resizebox{\textwidth}{!}{%
\begin{tabular}{l|cccc|cccc|cccc|cccc}
\toprule
\multirow{2}{*}{\textbf{Method}} & \multicolumn{4}{c|}{\textbf{Camelyon16 (HIS)}} & \multicolumn{4}{c|}{\textbf{SICAPv2}} & \multicolumn{4}{c|}{\textbf{NCT-CRC}} & \multicolumn{4}{c}{\textbf{BRACS}} \\
\cmidrule(lr){2-5} \cmidrule(lr){6-9} \cmidrule(lr){10-13} \cmidrule(lr){14-17}
 & AUC & AP & F1 & ACC & AUC & AP & F1 & ACC & AUC & AP & F1 & ACC & AUC & AP & F1 & ACC \\
\midrule
\multicolumn{17}{c}{\textit{\textbf{Setting A: General Domain Backbone (CLIP ViT-L/14)}}} \\
\midrule
CoOp & 73.12 & 71.93 & 71.16 & 64.40 & 80.05 & 82.11 & 73.33 & 69.45 & 77.90 & 57.92 & 62.21 & 68.77 & 60.36 & 68.73 & 75.39 & 60.90 \\
MedCLIP & 76.96 & 71.74 & 75.18 & 71.96 & 79.45 & 82.84 & 73.79 & 73.65 & 76.63 & 62.03 & 60.19 & 70.03 & 65.88 & 72.78 & 75.73 & 64.18 \\
DRA & 70.29 & 68.72 & 71.59 & 62.44 & 78.41 & 80.81 & 72.27 & 67.20 & 80.79 & 70.38 & 66.55 & 75.82 & 63.11 & 70.45 & 76.24 & 63.30 \\
WinCLIP & 72.87 & 62.82 & 74.71 & 68.40 & 73.13 & 74.51 & 70.23 & 64.65 & 75.86 & 55.18 & 61.01 & 65.81 & 56.16 & 64.63 & 75.07 & 60.10 \\
VAND & 77.41 & 73.24 & 74.07 & 70.21 & 83.15 & 85.73 & 76.03 & \underline{74.50} & 81.32 & 59.51 & 64.28 & 69.43 & 67.06 & 74.11 & 75.89 & 63.90 \\
MMA & 79.90 & 72.94 & \underline{78.15} & \underline{75.16} & 82.57 & 84.84 & 75.80 & 74.40 & 79.44 & 69.09 & 64.39 & 71.74 & 69.39 & 74.53 & 78.05 & 67.73 \\
MVFA &\underline{82.71} & 80.42 & 76.73 & 74.21 & \underline{84.75} & \underline{87.19} & \underline{77.02} & 74.25 & 82.32 & 61.10 & 65.17 & 70.03 & 70.33 & \underline{75.73} & 77.97 & 68.33 \\
MadCLIP & 80.60 & \underline{80.50} & 75.04 & 71.16 & 82.90 & 86.25 & 74.79 & 74.05 & \textbf{84.45} & \underline{72.31} & \underline{65.74} & \underline{72.18} & \underline{71.02} & 75.19 & \underline{78.17} & \underline{69.05} \\ 
\rowcolor{gray!10} \textbf{HAAF (Ours)} & \textbf{84.98} & \textbf{83.79} & \textbf{78.89} & \textbf{76.77} & \textbf{85.56} & \textbf{87.49} & \textbf{78.01} & \textbf{76.60} & \underline{83.95} & \textbf{73.32} & \textbf{66.33} & \textbf{74.40} & \textbf{71.89} & \textbf{75.84} & \textbf{78.34} & \textbf{69.21} \\
\midrule
\multicolumn{17}{c}{\textit{\textbf{Setting B: Pathology Foundation Backbone (CONCH ViT-L/14)}}} \\
\midrule
CoOp$^\dagger$ & 79.70 & 80.08 & 73.46 & 70.41 & 89.61 & 91.51 & 82.02 & 81.85 & \underline{88.02} & {83.82} & \underline{73.07} & \underline{84.95} & 78.62 & 84.17 & 78.70 & 71.32 \\
DRA$^\dagger$ & 78.23 & 77.96 & 73.78 & 71.11 & \underline{91.59} & \underline{93.51} & 84.59 & 85.30 & 87.20 & 79.24 & 71.54 & 82.14 & 67.84 & 76.71 & 75.06 & 60.08 \\
WinCLIP$^\dagger$ & 78.51 & 78.24 & 73.97 & 71.31 & 90.23 & 92.36 & 83.12 & 84.40 & 85.77 & 65.40 & 70.04 & 81.06 & 80.35 & 84.15 & 81.30 & 74.73 \\
VAND$^\dagger$ & 87.35 & 86.54 & 80.40 & 78.47 & 88.98 & 90.63 & 82.82 & 83.05 & 86.55 & 74.78 & 68.68 & 79.83 & 81.39 & \underline{84.98} & 81.81 & 76.05 \\
MMA$^\dagger$ & 86.09 & 84.87 & 79.76 & 77.67 & 87.62 & 89.71 & 79.41 & 77.75 & 86.59 & 78.63 & 71.30 & 82.29 & 80.52 & 84.02 & 81.07 & 73.94 \\
MVFA$^\dagger$ & \underline{88.51} & 87.65 & 81.29 & 79.37 & 91.22 & 93.02 & \underline{85.21} & \underline{85.75} & 86.75 & 76.09 & 68.63 & 79.73 & 81.04 & 84.81 & 81.41 & 74.54 \\
MadCLIP$^\dagger$ & 88.38 & \underline{88.63} & \underline{81.33} & \underline{79.82} & 86.17 & 89.26 & 77.06 & 78.65 & 87.98 & 81.11 & 67.76 & 80.16 & \underline{82.32} & 84.19 & \underline{82.70} & \underline{76.89} \\
\rowcolor{gray!10} \textbf{HAAF (Ours)} & \textbf{91.97} & \textbf{92.16} & \textbf{84.39} & \textbf{83.93} & \textbf{94.05} & \textbf{95.37} & \textbf{88.22} & \textbf{88.50} & \textbf{90.25} & \textbf{86.75} & \textbf{87.98} & \textbf{87.20} & \textbf{83.53} & \textbf{86.08} & \textbf{84.22} & \textbf{79.14} \\
\bottomrule
\end{tabular}%
}
\vspace{0mm}
\\
\raggedright \footnotesize{$^\dagger$ Indicates our re-implementation of the method using the CONCH backbone.}
\vspace{-5pt}
\end{table*}

\vspace{-3mm}
\subsubsection{\textbf{Superiority with General Backbone.}}
In the top block of Table~\ref{tab:main_results}, HAAF achieves competitive or superior performance compared to existing baselines using the standard CLIP backbone. Specifically, on the Camelyon16 (HIS) dataset, HAAF achieves an AUC of 84.98\%, surpassing the strong competitor MVFA (82.71\%) by 2.27\%. On the NCT-CRC benchmark, although MadCLIP shows a marginal advantage in AUC (84.45\% vs. 83.95\%), HAAF significantly outperforms it in Average Precision (73.32\% vs. 72.31\%) and F1-score.
This indicates that our method maintains a better precision-recall balance, which is often more critical in medical anomaly detection where false positives can increase clinical workload.

\vspace{-1mm}
\subsubsection{\textbf{Scalability with Pathology Foundation Models.}}
The bottom block of Table~\ref{tab:main_results} highlights the impact of switching to the pathology-specific CONCH backbone.
HAAF achieves new state-of-the-art results across all benchmarks, with AUC scores reaching 91.97\%, 94.05\%, 90.25\%, and 83.53\% on HIS, SICAP, CRC, and BRACS, respectively.
Crucially, comparing against MVFA equipped with the same CONCH backbone, HAAF maintains a clear lead (e.g., +3.46\% AUC on HIS).
This confirms that the performance gains stem from the proposed CLSA mechanism unlocking the fine-grained semantic potential of pathology foundation models, rather than solely from the stronger backbone.

\subsection{Few-Shot Robustness Analysis}
\label{sec:few_shot_analysis}

\begin{figure}[t]
  \setlength{\belowcaptionskip}{-10pt}
  \centering
   \includegraphics[width=\linewidth]{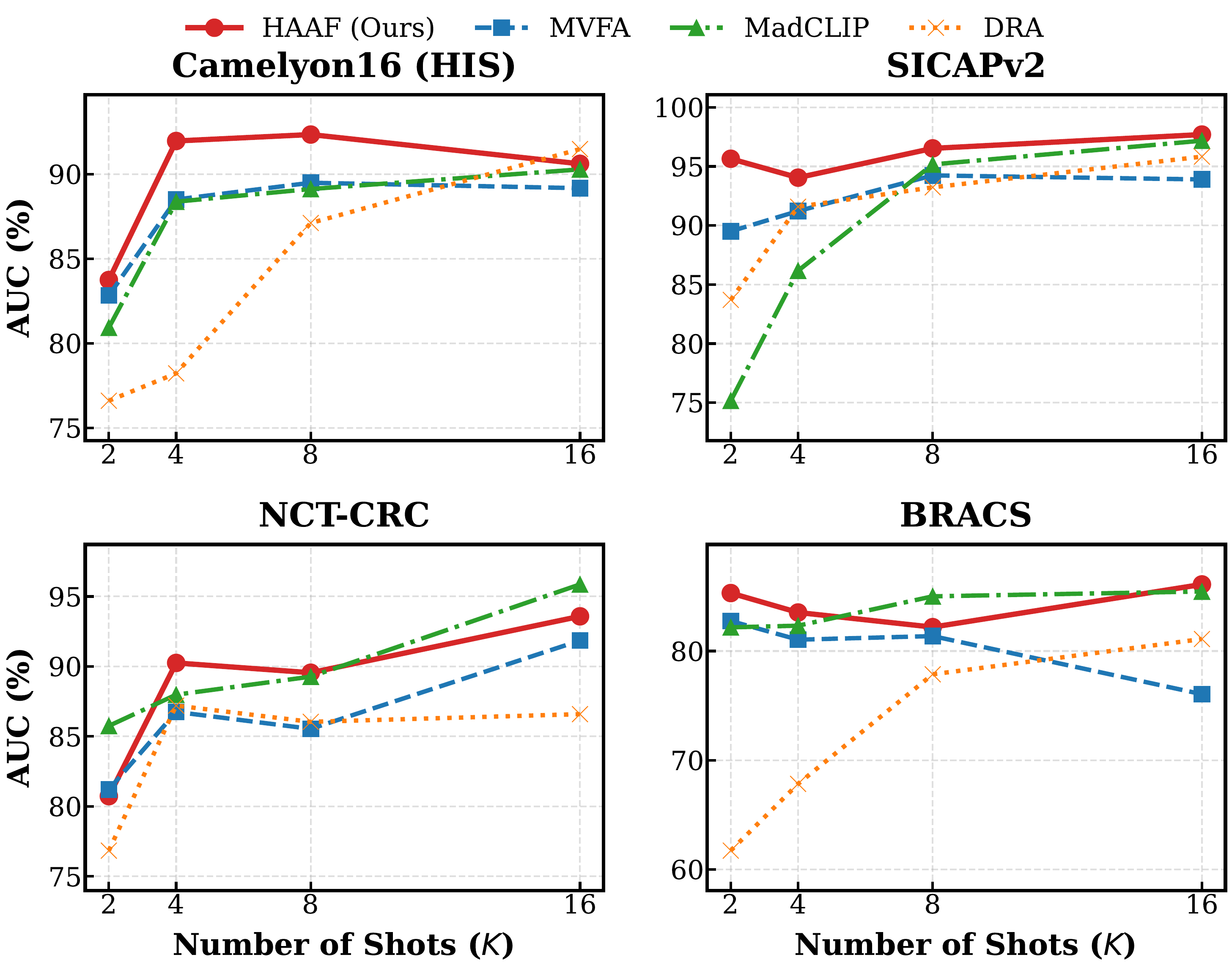} 
  \caption{Few-shot robustness analysis across four histopathology datasets.}
  \label{fig:few_shot}
  \vspace{-2mm}
\end{figure}

Having established the performance at 4-shot, we further evaluate data efficiency by comparing HAAF with MVFA, MadCLIP, and DRA across varying support set sizes ($K \in \{2, 4, 8, 16\}$) using the CONCH backbone (Figure~\ref{fig:few_shot}). Please refer to \textbf{Appendix~\ref{sec:few_shot_robustness}} for detailed numerical results corresponding to these plots.

\vspace{1mm}
\noindent\textbf{Performance Stability.}
HAAF consistently maintains a leading position. Notably, on \textbf{SICAPv2} and \textbf{Camelyon16}, HAAF exhibits dominance across all shot settings.
On \textbf{BRACS}, we observe that baseline methods (e.g., MVFA) suffer from performance degradation as the shot number increases (dropping to 76.05\% at 16-shot). We attribute this performance degradation in baselines to `Prototype Pollution'. As the support set grows ($K=16$), the inclusion of borderline cases (e.g., subtle Atypia in BRACS) introduces high variance into the class prototypes. Standard methods, which rely on simple averaging, are susceptible to these outliers. HAAF mitigates this through its dual-branch design: the semantic alignment branch acts as a dynamic filter, down-weighting non-representative visual features, thereby preserving the discriminative integrity of the prototypes even in the presence of noise.
In contrast, HAAF remains resilient (maintaining accuracy above 82\%), suggesting that our sequential cross-modal alignment effectively filters out semantic noise, ensuring that the prototypes remain discriminative even when the support set contains ``hard'' examples.

\subsection{Ablation Study}
\label{sec:ablation}

We conduct extensive ablation studies on the CONCH backbone (4-shot) to verify the contribution of each component (Table~\ref{tab:ablation}).

\begin{table}[t]
\centering
\caption{Ablation study of key components and interaction strategies on the CONCH backbone (4-shot). \textbf{I-Adapter}: Intra-modal Adapters; \textbf{Dual}: Dual-Branch Inference.}
\label{tab:ablation}
\setlength{\tabcolsep}{3pt} 
\resizebox{\columnwidth}{!}{%
\begin{tabular}{c|ccc|c|c}
\toprule
\textbf{Row} & \textbf{I-Adapter} & \textbf{CLSA Strategy} & \textbf{Dual} & \textbf{AUC (\%)} & \textbf{AP (\%)} \\
\midrule
1 & - & - & - & 70.97 & 62.26 \\ 
2 & \checkmark & - & - & 78.36 & 70.75 \\ 
3 & \checkmark & - & \checkmark & 83.79 & 82.97 \\ 
\midrule
4 & \checkmark & V $\to$ T (Only) & \checkmark & 87.95 & 89.34 \\ 
5 & \checkmark & T $\to$ V (Only) & \checkmark & 88.26 & 87.77 \\ 
\rowcolor{gray!10} 6 & \checkmark & \textbf{V $\to$ T $\to$ V (Seq.)} & \checkmark & \textbf{91.97} & \textbf{92.16} \\ 
\bottomrule
\end{tabular}
}
\vspace{-3mm}
\end{table}

\subsubsection{\textbf{Component Effectiveness.}}
Simply adding Intra-modal Adapters (Row 2) yields a significant gain over the baseline. Interestingly, introducing the Dual-Branch strategy without cross-modal alignment (Row 3) also improves performance, confirming that geometric prototypes provide a robust safety net.

\subsubsection{\textbf{Impact of Sequential Interaction Strategy.}}
Rows 4-6 investigate the design of CLSA. We observe that:
(1) \textbf{Unidirectional interaction} provides limited improvement.
(2) The superiority of the full sequential strategy (Row 6) confirms our core hypothesis: Interaction direction matters. Using only $V \to T$ (Context Injection) lacks spatially precise guidance for the final visual map.
Conversely, using only $T \to V$ (Semantic Guidance) forces the visual features to align with a generic, uncalibrated text prompt. Only the sequential $V \to T \to V$ design ensures guidance is driven by an content-specific semantic goal, creating a virtuous cycle of refinement.

\subsubsection{\textbf{Comparison with Standard PEFT Strategies.}}
To further verify that the superiority of HAAF stems from our specific Cross-Modal Linear Self-Alignment (CLSA) design rather than merely introducing additional learnable parameters, we compared HAAF against two mainstream Parameter-Efficient Fine-Tuning (PEFT) methods: \textbf{Standard Adapter}~\cite{houlsby2019parameter} and \textbf{LoRA}~\cite{hu2022lora}.

For a fair comparison, we integrated these distinct modules into the same four stages of the CONCH vision encoder, keeping the frozen backbone identical.
As shown in Table~\ref{tab:peft_comparison}, while standard Adapter and LoRA provide improvements, they still fall short compared to HAAF.
Specifically, HAAF outperforms the stronger competitor (Adapter or LoRA) by a significant margin (e.g., \textbf{+6.93\%} on Camelyon16).
This is attributed to the fact that generic PEFT methods~\cite{liu2024moe} operate solely within the visual modality, whereas HAAF explicitly utilizes text embeddings to dynamically calibrate visual features layer-by-layer.

\begin{table}[h]
\centering
\vspace{-1mm}
\caption{\textbf{Comparison with standard PEFT methods (4-shot).}}
\label{tab:peft_comparison}
\setlength{\tabcolsep}{2pt}
\resizebox{\columnwidth}{!}{%
\begin{tabular}{l|c|c|c|c}
\toprule
\textbf{Method} & \textbf{Camelyon16} & \textbf{SICAPv2} & \textbf{NCT-CRC} & \textbf{BRACS} \\
\midrule
CONCH + Adapter & 85.04 & 88.83 & 86.63 & 81.15 \\
CONCH + LoRA    & 84.90 & 90.72 & 87.42 & 80.96 \\
\rowcolor{gray!10} \textbf{HAAF (Ours)} & \textbf{91.97} & \textbf{94.05} & \textbf{90.25} & \textbf{83.53} \\
\bottomrule
\end{tabular}
\vspace{-1em}
}
\end{table}

\vspace{1mm}
\subsubsection{\textbf{Impact of Hierarchical Adaptation Stages.}}
To validate the necessity of our multi-stage design, we evaluated the performance when applying the HAAF module to single stages individually versus the proposed multi-stage fusion (Table~\ref{tab:layer_analysis}).
We observe a clear trend:
(1) \textbf{Single-stage limitations:} Applying the adapter to any single stage alone yields sub-optimal results. While deeper layers (e.g., Stage 4) generally perform better, they fail to capture the full spectrum of pathological features.
(2) \textbf{Synergy of Multi-stage Fusion:} The combined strategy (``All Stages'') consistently achieves the highest AUC. For instance, on the fine-grained SICAPv2 dataset, the multi-stage fusion outperforms the best single-stage counterpart (Stage 4) by \textbf{1.07\%}.

\begin{table}[!t]
\centering
\caption{\textbf{Layer-wise ablation analysis (4-shot).}}
\label{tab:layer_analysis}
\setlength{\tabcolsep}{3pt} 
\resizebox{\columnwidth}{!}{%
\begin{tabular}{l|c|c|c|c}
\toprule
\textbf{Applied Stage} & \textbf{Camelyon16} & \textbf{SICAPv2} & \textbf{NCT-CRC} & \textbf{BRACS} \\
\midrule
Stage 1 (Shallow) & 79.97 & 84.82 & 79.03 & 74.16 \\
Stage 2           & 84.17 & 89.56 & 83.97 & 77.53 \\
Stage 3           & 87.63 & 91.67 & 86.01 & 79.89 \\
Stage 4 (Deep)    & 90.53 & 92.98 & 88.43 & 81.31 \\
\midrule
\rowcolor{gray!10} \textbf{All Stages (Ours)} & \textbf{91.97} & \textbf{94.05} & \textbf{90.25} & \textbf{83.53} \\
\bottomrule
\end{tabular}
}
\vspace{-2em}
\end{table}

\subsection{Visualization and Interpretability}
\label{sec:vis}

To verify whether HAAF correctly identifies pathological regions beyond image-level metrics, we conducted a pixel-level visualization on the SICAP dataset.
\textit{Setup:} We attached a lightweight auxiliary Anomaly Segmentation head to the frozen encoder.
\textbf{Crucially, the available pixel-level masks were utilized solely to train this auxiliary head for visualization and were strictly excluded from the main few-shot classification training}, ensuring the fairness of the setting.

\begin{figure*}[t]
    \centering
    \includegraphics[width=0.98\textwidth]{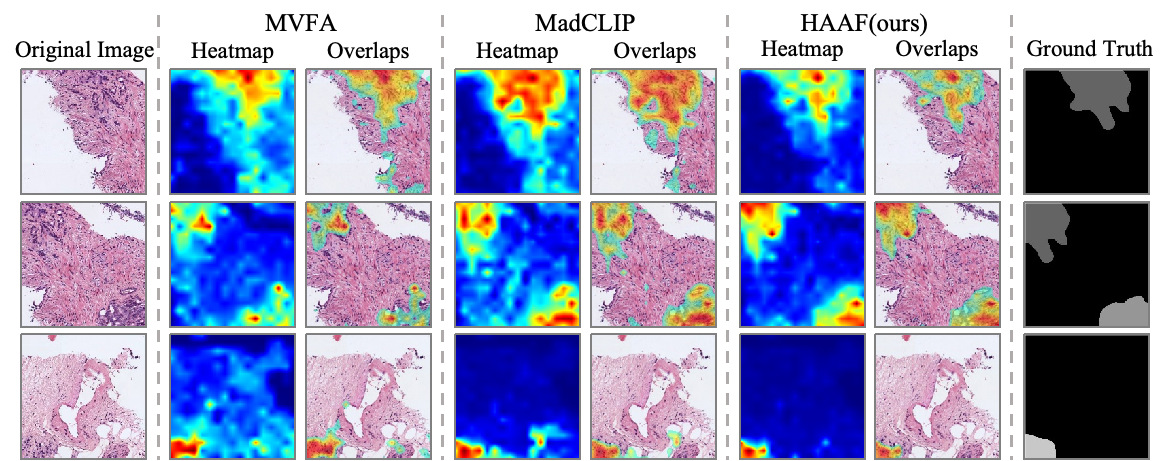} 
    \caption{\textbf{Qualitative visualization on SICAP.} 
    From left to right: Original H\&E, MVFA, MadCLIP, HAAF, and Ground Truth. Red indicates high anomaly scores. 
    Compared to baselines, HAAF demonstrates superior localization of minute lesions (Row 1) and boundary alignment (Row 2), verifying its fine-grained interpretability.} 
    \label{fig:vis}
    \vspace{-3mm}
\end{figure*}

Figure~\ref{fig:vis} compares our method with MVFA and MadCLIP. The visualization reveals three key observations:
\begin{itemize}[leftmargin=*]
    \item \textbf{Precise Localization:} For samples with minute lesions (e.g., first row), competitor methods like MadCLIP often fail to localize the target. In contrast, HAAF accurately pinpoints the focal area.
    \item \textbf{Boundary Alignment:} For larger tumor regions (e.g., third row), our heatmaps demonstrate superior alignment with the Ground Truth, closely following the morphological structure.
    \item \textbf{Noise Suppression:} MVFA tends to generate scattered noise in benign regions (background). Our method effectively suppresses these background activations, resulting in a cleaner and more reliable anomaly map.
\end{itemize}
These results confirm that HAAF possesses strong semantic grounding at the pixel level.

\vspace{-1mm}
\section{Related Works}
\textbf{Vision–Language Foundation Models.}
Vision-language models (VLMs) like CLIP \cite{Radford2021LearningTV} offer a transformative approach for anomaly detection (AD) by aligning visual and textual concepts. Early works \cite{Jeong2023WinCLIPZA, Li2024PromptADLP, Zhu2024TowardGA} demonstrated effective zero-shot defect localization, a capability extended to medical imaging to alleviate annotation needs \cite{Shiri2025MadCLIPFM, Cao2023PersonalizingVM}. This foundation has spurred exploration of novel geometries \cite{GonzlezJimnez2025IsHS} and generative backbones \cite{Wang2025PathologyInformedLD}. The frontier now focuses on explainability, leveraging universal adaptation \cite{Ma2025AACLIPEZ, Gao2025AdaptCLIPAC} and LLM-powered reasoning \cite{Jin2025LogicADEA, Gu2023AnomalyGPTDI} to move from black-box detection to interpretable analysis. However, directly transferring these general-domain paradigms to histopathology remains non-trivial due to the significant semantic gap and the need for fine-grained expert knowledge.

\vspace{1mm}
\noindent\textbf{Zero-/Few-Shot Anomaly Detection.}
Traditional anomaly detection (AD) is shifting from methods reliant on reconstruction errors or density estimation \cite{Roth2021TowardsTR, Liu2023SimpleNetAS} towards leveraging pre-trained foundation models. Among these, VLMs have emerged as a dominant architecture. Early CLIP-based works \cite{Jeong2023WinCLIPZA} established sliding-window inference, later refined by techniques employing semantic concatenation and in-context learning \cite{Li2024PromptADLP, Zhu2024TowardGA}. To bridge domain gaps, subsequent research introduced universal adapters \cite{Gao2025AdaptCLIPAC} and anomaly-aware prompts \cite{Ma2025AACLIPEZ}. Beyond classification, recent efforts focus on explainable and pixel-precise localization, integrating foundational segmentation models \cite{Cao2023PersonalizingVM}, LLMs for textual reasoning \cite{Gu2023AnomalyGPTDI}, and diffusion models for reconstruction \cite{He2023DiADAD}, with unified benchmarks \cite{You2022AUM} establishing rigorous evaluations. Despite this progress, most VLM-based AD methods still struggle to distinguish visually similar normal variants from subtle lesions without explicit, instance-aware semantic guidance.

\vspace{1mm}
\noindent\textbf{Medical Anomaly Detection.}
Medical anomaly detection faces challenges from subtle pathologies, data scarcity, and domain shifts. Modern approaches leverage foundation models to mitigate these issues. VLMs are enhanced through domain-specific knowledge infusion, as seen in few-shot adapters \cite{Shiri2025MadCLIPFM}, pathology-informed guidance \cite{Song2025NormalAA, Zhao2021SpikingCA}, and foundational segmentation models like MedSAM \cite{Ma2023SegmentAI} for pixel-precise localization. Concurrently, generative and self-supervised strategies utilize pathology-informed diffusion \cite{Wang2025PathologyInformedLD} and specialized representation learning \cite{Hou2024ASF, Mekala2023SELFOODSO, Tuli2022TranADDT}, alongside architectures such as MedFormer \cite{Xia2025MedFormerHM}, yet effective synergy between V-L reasoning and few-shot learning remains underexplored.
\vspace{-5mm} 
\section{Conclusion}

We presented \textbf{HAAF}, a data-efficient framework to steer Vision-Language foundation models toward fine-grained pathology anomaly detection. The core innovation lies in the \textit{Cross-Level Scaled Alignment (CLSA)}, which shifts the paradigm from static, parallel feature fusion to a dynamic, sequential calibration process. By enabling visual features to contextualize text prompts prior to semantic guidance, HAAF overcomes the limitations of generic representations. Coupled with a robust dual-branch scoring strategy, our framework demonstrates remarkable versatility across diverse anatomical regions and backbones. Our results conclusively show that hierarchical, sequential alignment is essential for adapting both general (CLIP) and medical (CONCH) foundation models to the rigorous demands of ROI-level diagnosis, paving the way for accessible automated pathology systems.
\vspace{-3mm}
\begin{acks}
This research was supported by the National Science and Technology Major Project (Grant No.~2025ZD0544802), the Key Research and Development Program of Shaanxi Province (Grant No.~2024SF-GJHX-32), the Key Research and Development Program of Ningxia Hui Autonomous Region (Grant No.~2023BEG02023), the Noncommunicable Chronic Diseases–National Science and Technology Major Project (Grant No.~2024ZD0527700), the project ``Research on Key Technologies for Full-Chain Intelligent Pathological Diagnosis'' (Grant No.~HX202440) of The First Affiliated Hospital of Xi'an Jiaotong University, the National Natural Science Foundation of China (No.~62506291), and the XJTU Research Fund for AI Science (No.~2025YXYC004). Z.G. was supported by GE HealthCare.
\end{acks}

\bibliographystyle{ACM-Reference-Format}
\bibliography{main}

\clearpage
\appendix
\section{Experimental Settings}

\subsection{Detailed Dataset Descriptions}
\label{sec:app_datasets}

In this section, we provide detailed configurations for the four histopathology datasets used in our benchmark. For all datasets, we reformulated the original tasks into a one-vs-all anomaly detection setting, where healthy or benign tissues are defined as \textit{Normal}, and cancerous or pre-cancerous lesions are defined as \textit{Abnormal}.

\noindent\textbf{1. Camelyon16 (Breast Metastasis)~\cite{Camelyon16}.}
We adopt the standardized anomaly detection setting established in the \textbf{BMAD benchmark}~\cite{bao2024bmad}. Derived from 400 Whole-Slide Images (WSIs) of sentinel lymph nodes, this benchmark provides curated, non-overlapping patches at 40$\times$ magnification with a resolution of $256 \times 256$ pixels, \textbf{specifically designed to evaluate the detection of small metastatic foci within complex lymph node backgrounds.}
\begin{itemize}[leftmargin=*]
\item \textbf{Statistics:} Following the BMAD protocol, we construct the few-shot support set by sampling from the official pool of benign patches. The performance is rigorously evaluated on a fixed test set comprising \textbf{2,000} images, consisting of a balanced collection of \textbf{1,000} normal and \textbf{1,000} tumor patches derived from 115 testing slides.
\end{itemize}

\noindent\textbf{2. SICAPv2 (Prostate Gleason Grading)~\cite{SICAPv2}.}
This dataset consists of 18,783 histological images at 10$\times$ magnification derived from 155 patients. It was originally designed for Gleason Grading (GG), a critical prognostic indicator for prostate cancer. The dataset includes four grades: non-cancerous (GG0), and cancerous grades GG3 (atrophic/dense glands), GG4 (cribriform/fused glands), and GG5 (solid/single cells).
We map these grades to binary classes as follows:
\begin{itemize}[leftmargin=*]
\item \textbf{Normal Class:} Consists of patches annotated as GG0.
\item \textbf{Abnormal Class:} Aggregates all cancerous grades (GG3, GG4, and GG5).
\item \textbf{Statistics:} To ensure a robust evaluation, we constructed a fixed test set by randomly sampling \textbf{1,000} normal images and \textbf{1,000} abnormal images from the dataset, independent of the few-shot support set.
\end{itemize}

\noindent\textbf{3. NCT-CRC (Colorectal Cancer)~\cite{kather2019predictingCRC}.}
The NCT-CRC-HE-100K dataset provides a large-scale collection of non-overlapping histology image tiles ($224 \times 224$ pixels) from colorectal cancer patients. Images are acquired at 0.5 microns per pixel (MPP) and color-normalized using Macenko's method. The dataset originally covers 9 tissue classes.
We construct a challenging anomaly detection setting with high intra-class variance, \textbf{requiring the model to generalize across diverse healthy phenotypes (e.g., muscle, mucosa) while identifying tumor characteristics:}
\begin{itemize}[leftmargin=*]
\item \textbf{Normal Class:} Includes Adipose (ADI), Lymphocytes (LYM), Mucus (MUC), Smooth Muscle (MUS), and Normal Colon Mucosa (NORM).
\item \textbf{Abnormal Class:} Includes Colorectal Adenocarcinoma Epithelium (TUM), Cancer-associated Stroma (STR), and Debris (DEB).
\item \textbf{Excluded:} Background (BACK) patches are removed.
\item \textbf{Statistics:} The few-shot support set is sampled from the available normal pool. The evaluation is performed on a test set comprising \textbf{4,324} normal patches and \textbf{1,973} abnormal patches.
\end{itemize}

\noindent\textbf{4. BRACS (Breast Carcinoma Subtyping)~\cite{brancati2022bracs}.}
The BReAst Carcinoma Subtyping (BRACS) dataset contains Regions of Interest (ROIs) representing a spectrum of breast lesions. We map the original 7 subtypes into a binary split based on clinical malignancy:
\begin{itemize}[leftmargin=*]
\item \textbf{Normal Class:} Includes Normal (N), Pathological Benign (PB), and Usual Ductal Hyperplasia (UDH).
\item \textbf{Abnormal Class:} Includes pre-cancerous and cancerous lesions, specifically Flat Epithelial Atypia (FEA), Atypical Ductal Hyperplasia (ADH), Ductal Carcinoma In Situ (DCIS), and Invasive Carcinoma (IC). \textbf{Distinguishing these borderline pre-cancerous lesions from benign mimics involves capturing subtle nuclear and architectural atypia.}
\item \textbf{Statistics:} The few-shot support set is sampled from the normal pool. The evaluation is performed on a test set comprising \textbf{1,460} normal images and \textbf{2,197} abnormal images.
\end{itemize}

\vspace{-2mm}
\subsection{Details of Baseline Methods}
\label{sec:app_baselines}

To evaluate the proposed HAAF framework comprehensively, we compare it against a diverse set of baselines ranging from general foundation models to state-of-the-art (SOTA) few-shot anomaly detection methods.

\vspace{2pt}
\noindent\textbf{1. General Vision-Language Adaptation.}
\textbf{CLIP}~\cite{radford2021learningCLIP} serves as the fundamental backbone. We utilize its pre-trained visual and textual encoders with standard hand-crafted prompt templates (e.g., ``a photo of a [CLASS]'') to generate static classifier weights, serving as a baseline to evaluate the benefit of learnable adaptation.
\textbf{CoOp}~\cite{zhou2022learningCOOP} addresses the rigidity of hand-crafted prompts by introducing continuous learnable context vectors. Instead of manually designing templates, CoOp optimizes these vectors in the embedding space using the support set, effectively automating the prompt engineering process to align with the target distribution.

\vspace{2pt}
\noindent\textbf{2. Medical-Specific Vision-Language Models.}
\textbf{MedCLIP}~\cite{hua2024medicalclip} bridges the domain gap by pre-training on large-scale medical image-text pairs (e.g., ROCO). It utilizes a decoupled learning strategy to leverage unmatched data. We evaluate MedCLIP to assess whether domain-specific pre-training alone provides sufficient discriminability for fine-grained anomaly detection without specialized adaptation modules.

\vspace{2pt}
\noindent\textbf{3. Anomaly Detection Baselines.}
\textbf{WinCLIP}~\cite{jeong2023winclip} enhances CLIP with a compositional prompt ensemble and multi-scale window-based aggregation to capture local anomalies often overlooked by global embeddings.
\textbf{VAND}~\cite{vand2023} adapts CLIP by refining V-L alignment to focus on anomaly-related descriptions, utilizing mapping strategies to improve the detection of deviations.
\textbf{DRA}~\cite{ding2022catchingDRA} disentangles normal patterns from anomalous deviations, effectively improving robustness in few-shot scenarios by preventing overfitting to background noise in the support set.

\vspace{2pt}
\noindent\textbf{4. Recent Multi-Modal AD SOTA.}
\textbf{MMA}~\cite{yang2024mma} introduces lightweight adapters to both visual and textual branches to fine-tune pre-trained features. However, similar to other approaches, it largely processes modalities independently before the final alignment step.
\textbf{MVFA}~\cite{huang2024mvfa} enhances detection by aggregating features from multiple views or scales. It employs a parallel fusion strategy to combine these features with text embeddings, though it lacks a mechanism to strictly calibrate text prompts based on instance-specific visual inputs.
\textbf{MadCLIP}~\cite{Shiri2025MadCLIPFM} represents the current state-of-the-art, employing a multi-scale adaptation strategy and refined prompt engineering. We compare against it to demonstrate the superiority of HAAF's sequential interaction mechanism over existing parallel adaptation schemes.

\begin{table*}[t!] 
\centering
\caption{\textbf{Comprehensive comparison of few-shot performance across varying support set sizes ($K \in \{2, 4, 8, 16\}$).} All methods use the CONCH backbone. We report AUC, AP, F1-score, and Accuracy (\%). The best results for each setting are highlighted in \textbf{bold}.}
\label{tab:full_few_shot}
\setlength{\tabcolsep}{1.8pt} 
\renewcommand{\arraystretch}{1.0} 
\resizebox{\textwidth}{!}{%
\begin{tabular}{l|c|cccc|cccc|cccc|cccc}
\toprule
\multirow{2}{*}{\textbf{Dataset}} & \multirow{2}{*}{\textbf{Shot}} & \multicolumn{4}{c|}{\textbf{HAAF (Ours)}} & \multicolumn{4}{c|}{\textbf{MVFA}~\cite{huang2024mvfa}} & \multicolumn{4}{c|}{\textbf{MadCLIP}~\cite{Shiri2025MadCLIPFM}} & \multicolumn{4}{c}{\textbf{DRA}~\cite{ding2022catchingDRA}} \\
\cmidrule(lr){3-6} \cmidrule(lr){7-10} \cmidrule(lr){11-14} \cmidrule(lr){15-18}
 &  & AUC & AP & F1 & ACC & AUC & AP & F1 & ACC & AUC & AP & F1 & ACC & AUC & AP & F1 & ACC \\
\midrule
\multirow{4}{*}{\textbf{HIS}} 
& 2 & \textbf{83.75} & \textbf{83.05} & \textbf{77.31} & \textbf{76.36} & 82.84 & 74.75 & 79.84 & 77.87 & 80.93 & 80.59 & 74.59 & 71.51 & 76.61 & 77.56 & 71.40 & 67.70 \\
& 4 & \textbf{91.97} & \textbf{92.16} & \textbf{84.39} & \textbf{83.93} & 88.51 & 87.65 & 81.29 & 79.37 & 88.38 & 88.63 & 81.33 & 79.82 & 78.23 & 77.96 & 73.78 & 71.11 \\
& 8 & \textbf{92.35} & \textbf{93.59} & \textbf{85.51} & \textbf{85.98} & 89.50 & 90.45 & 81.17 & 80.97 & 89.13 & 90.72 & 81.19 & 81.67 & 87.12 & 87.05 & 80.58 & 79.27 \\
& 16 & \textbf{90.62} & \textbf{91.30} & \textbf{84.63} & \textbf{84.63} & 89.18 & 90.70 & 81.64 & 81.37 & 90.29 & 90.62 & 82.95 & 82.07 & 91.49 & 92.29 & 84.34 & 84.03 \\
\midrule
\multirow{4}{*}{\textbf{SICAP}} 
& 2 & \textbf{95.65} & \textbf{96.13} & \textbf{89.27} & \textbf{89.10} & 89.61 & 91.51 & 82.02 & 81.85 & 75.16 & 77.01 & 71.67 & 64.55 & 83.70 & 85.74 & 76.42 & 76.15 \\
& 4 & \textbf{94.05} & \textbf{95.37} & \textbf{88.22} & \textbf{88.50} & 91.22 & 93.02 & 85.21 & 85.75 & 86.17 & 89.26 & 77.06 & 78.65 & 91.59 & 93.51 & 84.59 & 85.30 \\
& 8 & \textbf{96.53} & \textbf{97.15} & \textbf{91.15} & \textbf{91.15} & 94.24 & 94.98 & 89.41 & 89.85 & 95.17 & 96.33 & 89.60 & 89.75 & 93.23 & 94.15 & 86.08 & 86.05 \\
& 16 & \textbf{97.70} & \textbf{98.17} & \textbf{93.84} & \textbf{93.95} & 93.90 & 94.71 & 87.26 & 87.40 & 97.18 & 97.60 & 92.54 & 92.50 & 95.83 & 96.71 & 89.83 & 90.30 \\
\midrule
\multirow{4}{*}{\textbf{CRC}} 
& 2 & 80.73 & 68.65 & 68.89 & \textbf{80.45} & 81.21 & 67.41 & 66.36 & 78.96 & \textbf{85.75} & \textbf{77.77} & 66.00 & 77.34 & 76.84 & 61.18 & 60.24 & 73.62 \\
& 4 & \textbf{90.25} & \textbf{84.61} & \textbf{73.25} & \textbf{84.48} & 86.75 & 76.09 & 68.63 & 79.73 & 87.98 & 81.11 & 67.76 & 80.16 & 87.20 & 79.24 & 71.54 & 82.14 \\
& 8 & \textbf{89.55} & \textbf{84.15} & \textbf{74.57} & \textbf{85.06} & 85.54 & 71.28 & 68.80 & 79.10 & 89.27 & 82.85 & 70.73 & 81.83 & 86.04 & 77.55 & 70.86 & 82.41 \\
& 16 & 93.58 & 88.22 & 83.94 & \textbf{89.55} & 91.85 & 86.23 & 80.21 & 87.03 & \textbf{95.86} & \textbf{92.54} & \textbf{83.92} & \textbf{89.74} & 86.59 & 78.63 & 71.30 & 82.29 \\
\midrule
\multirow{4}{*}{\textbf{BRACS}} 
& 2 & \textbf{85.31} & \textbf{89.55} & \textbf{82.71} & 77.28 & 82.75 & 85.59 & \textbf{82.92} & 77.06 & 82.17 & 84.00 & 82.55 & \textbf{77.06} & 61.73 & 71.80 & 75.06 & 60.08 \\
& 4 & \textbf{83.53} & \textbf{86.08} & \textbf{84.22} & \textbf{79.14} & 81.04 & 84.81 & 81.41 & 74.54 & 82.32 & 84.19 & 82.70 & 76.89 & 67.84 & 76.71 & 75.06 & 60.08 \\
& 8 & 82.19 & 86.84 & 80.97 & 75.53 & 81.37 & 85.88 & 80.70 & 74.21 & \textbf{85.01} & \textbf{87.83} & \textbf{85.81} & \textbf{80.80} & 77.87 & 83.72 & 78.58 & 70.06 \\
& 16 & \textbf{86.10} & \textbf{89.19} & \textbf{82.89} & 78.01 & 76.05 & 73.96 & 82.26 & 76.59 & 85.45 & 88.59 & \textbf{84.40} & \textbf{79.30} & 81.10 & 86.48 & 80.50 & 72.87 \\
\bottomrule
\end{tabular}%
}
\vspace{-1em}
\end{table*}

\section{Supplementary Experiments}
\subsection{Comprehensive Few-Shot Robustness Analysis}
\label{sec:few_shot_robustness}

To further validate the data efficiency and stability of HAAF, we conducted a comprehensive evaluation across four datasets under varying few-shot settings, specifically $K \in \{2, 4, 8, 16\}$. We compared HAAF with three competitive baselines: MVFA~\cite{huang2024mvfa}, MadCLIP~\cite{Shiri2025MadCLIPFM}, and DRA~\cite{ding2022catchingDRA}, all implemented using the same CONCH backbone. The detailed quantitative results are reported in Table~\ref{tab:full_few_shot}.

\vspace{1mm}
\noindent\textbf{Performance at Extreme Low-Shot Regimes ($K=2$).}
HAAF demonstrates superior adaptation capabilities when training examples are extremely scarce. As shown in Table~\ref{tab:full_few_shot}, on the Camelyon16 and SICAP datasets, HAAF achieves 2-shot AUC scores of \textbf{83.75\%} and \textbf{95.65\%}, respectively, significantly outperforming the second-best methods. This indicates that our \textit{Cross-Level Scaled Alignment (CLSA)} mechanism can effectively extract discriminative features even with minimal reference samples.

\vspace{1mm}
\noindent\textbf{Robustness against Prototype Pollution ($K=16$).}
On BRACS, baselines like MVFA degrade significantly at 16-shot (AUC 82.75\%$\to$76.05\%) due to ``Prototype Pollution'' from noisy support samples. In contrast, HAAF remains resilient (\textbf{86.10\%} AUC), confirming that our sequential calibration effectively filters semantic noise to preserve prototype integrity.

\vspace{1mm}
\noindent\textbf{Consistency in Precision-Recall Balance.}
HAAF consistently secures leading AP scores (e.g., $>95\%$ on SICAP), demonstrating that our sequential interaction minimizes false positives and ensures reliable diagnosis in low-data regimes where baselines struggle.

\subsection{Hyperparameter Sensitivity}
\label{sec:app_sensitivity}

\begin{figure}[h]
  \centering
  \includegraphics[width=0.85\linewidth]{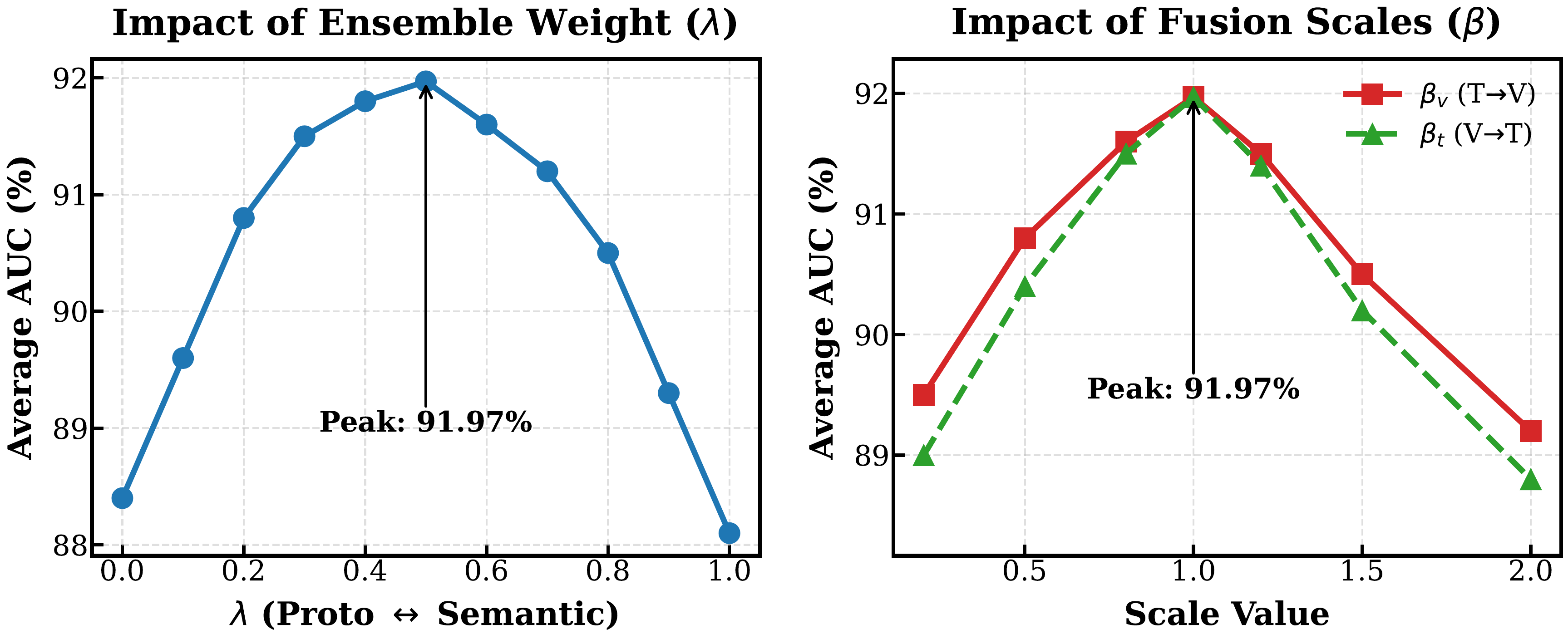} 
  \caption{\textbf{Sensitivity analysis.} (Left) Impact of ensemble weight $\lambda$. (Right) Impact of CLSA fusion scales.}
  \label{fig:sensitivity}
\end{figure}

We analyze the sensitivity of the ensemble weight $\lambda$ and CLSA fusion scales ($\beta$) using the CONCH backbone (Figure~\ref{fig:sensitivity}).

\noindent\textbf{Impact of Ensemble Weight ($\lambda$).}
Performance peaks around $\lambda = 0.5$, forming an inverted U-shaped curve. This confirms the complementary nature of our design: geometric prototypes provide stability against outliers, while semantic alignment offers discriminative guidance for fine-grained details, necessitating a balanced integration.

\noindent\textbf{Impact of Fusion Scales ($\beta$).}
The performance curve peaks at $\beta=1.0$. Reducing the scales ($\beta < 0.5$) leads to a drop in AUC due to insufficient adaptation strength. Conversely, excessively large scales ($\beta=2.0$) degrade performance by distorting the well-learned pre-trained foundation features, suggesting that a moderate fusion intensity is optimal for preserving feature integrity.


\end{sloppypar}
\end{document}